\documentclass{article}
\usepackage[utf8]{inputenc}
\usepackage{amsfonts}
\usepackage{spconf,amsmath,graphicx}
\usepackage{xcolor}
\usepackage{subfigure}
\usepackage[T1]{fontenc}
\usepackage{csquotes}
\usepackage{float}
\usepackage{bm}

\newtheorem{thm}{Theorem}[section]

\newtheorem{prop}[thm]{Proposition}

\name{Aniket Pramanik, Mathews Jacob}

\address{The University of Iowa, Iowa City, USA}
\begin{document}

\title{Improved Model based Deep Learning using Monotone Operator Learning (MOL)}

\maketitle

\begin{abstract}
Model-based deep learning (MoDL) algorithms that rely on unrolling are emerging as powerful tools for image recovery.  In this work, we introduce a novel monotone operator learning framework to overcome some of the challenges associated with current unrolled frameworks, including high memory cost, lack of guarantees on robustness to perturbations, and low interpretability. Unlike current unrolled architectures that use finite number of iterations, we use the deep equilibrium (DEQ) framework to iterate the algorithm to convergence and to evaluate the gradient of the convolutional neural network blocks using Jacobian iterations. This approach significantly reduces the memory demand, facilitating the extension of MoDL algorithms to high dimensional problems. We constrain the CNN to be a monotone operator, which allows us to introduce algorithms with guaranteed convergence properties and robustness guarantees. We demonstrate the utility of the proposed scheme in the context of parallel MRI.
\end{abstract}

\section{Introduction}

The recovery of biomedical images from few measurements using compressive sensing (CS) algorithm has received considerable attention in the last decade, with wide ranging applications including MRI and microscopy. 
In the recent years, deep learning methods have been revolutionizing image recovery. Model based (MoDL) architectures, which  unroll an iterative CS algorithm and replace the proximal mapping blocks by convolutional neural network (CNN) modules have been introduced by several authors including \cite{gregor2010learning,aggarwal2018modl}. While these schemes offer good performance and a significant reduction in training data demand, they have some limitations that restrict their utility. The main challenge in the training of these networks is the high memory demand. In particular, the images and features at each unrolling step need to be stored during back-propagation, which severely limits the number of unrolling steps in multi-dimensional applications. 
Another challenge is the interpretability of these models. Eventhough they are motivated by iterative algorithms, the limited number of unrolling steps often cannot guarantee the convergence of the algorithms to the minimum of any specific cost function. It is desirable to learn MoDL frameworks that inherit the desirable properties of classical convex optimization schemes.

Building upon deep equilibrium (DEQ) schemes that iterate a single layer \cite{deq} to convergence, a DEQ  architecture was recently introduced for image recovery \cite{gilton2021deep}. Assuming the network blocks to be non-expansive, the forward iterations are iterated until convergence \cite{gilton2021deep}. Instead of relying on unrolling for backpropagation, DEQ schemes rely on Jacobian backward iterations to compute the gradient of the CNN block. The main benefit of this approach is the quite significant reduction in the memory demand. 
Because the Lipshitz constant of the CNN block should be less than one for the forward and backward iterations to converge, spectral normalization was used in \cite{gilton2021deep}. 
We note that the Lipshitz constant computed using spectral methods of each layer is a very conservative upper bound. Our experiments in the the parallel MRI context show that the use of spectral normalization results in networks with limited expressive power, which translates to poor performance. Another challenge is that many algorithms such as ADMM and proximal gradients do not come with convergence guarantees in the MoDL setting when the data consistency (DC) term is not strictly convex (i.e. rank deficient forward model) \cite{gilton2021deep,ryu2019plug}.

In this work, we introduce a novel DEQ framework for inverse problems, where we focus on the learning of a deep monotone operator \cite{winston2020monotone}; we term this approach as monotone operator learning (MOL). We note that the derivative of convex priors are monotone operators. Even though the converse is not true, monotone operators can offer several benefits in the DEQ context, including guaranteed convergence and robustness to input perturbations. We introduce a forward backward algorithm, which uses the monotone operator. 
The proposed forward backward algorithm comes with convergence guarantees, provided the operator is strictly monotone. 
We show that any general residual network of the form $(I-H)$ involving a deep-network $H$ is strictly monotone, provided the Lipshitz constant of $H$ is less than unity. Because spectral normalization severely restricts the representation power, we propose an additional Lipshitz regularization term to the training loss. 
Our experiments show that this approach can offer significantly improved results over spectral normalized networks. 

We demonstrate the utility of the proposed scheme in the context of parallel MRI. Our results show that the proposed scheme can offer results that are comparable to unrolled networks, albeit with quite significant reduction in memory demand. This work paves the way for the extension of model based deep learning schemes to multi-dimensional imaging (e.g. 3D or 3D+time) applications.

\vspace{-1em}
\section{Methods}
\vspace{-1em}\subsection{Review of Deep Equilibrium Models}
We consider the recovery of an image $f:\mathbb R^2 \rightarrow \mathbb C$ from its undersampled linear measurements $\mathbf b = \mathcal A(f)+\mathbf n$, where $\mathcal A$ is a rank-deficient linear operator with maximum eigen value of 1 and $\mathbf n$ is noise. 
CS methods pose the recovery as an optimization problem with a convex prior
\vspace{-1em}
\begin{equation}\label{cvx}
f^* = \arg \min_f ~\underbrace{\frac{\lambda}{2}~\|\mathcal A(f)-\mathbf b\|^2+  \varphi(f)}_{C(f)}
\end{equation} 
\vspace{-1em}

Since $\varphi$ is often non-differentiable, iterative algorithms that rely on the proximal mapping of $\varphi$ 
 were introduced. 

In the recent years, 
several authors have introduced model-based deep learning (MoDL) algorithms, where an iterative compressed sensing algorithm is unrolled and the proximal operators are replaced with off-the-shelf denoisers or learned CNN modules. While these strategies are powerful, they suffer from high memory demand and low interpretability. The deep equilibrium algorithm was introduced to reduce the memory demand of MoDL algorithms. When $\varphi(f)$ is differentiable, the minimization of \eqref{cvx} satisfies the equation
\begin{equation}\label{fixedpoint}
\nabla_f~ C = \lambda \underbrace{A^T(Af-b)}_{G(f)} + \underbrace{\partial \varphi (f)}_{F(f)} = 0
\end{equation}
\vspace{-1em}

Gilton et al. \cite{gilton2021deep}, used iterative gradient descent $f_{n+1}= f_n - \gamma \nabla C(f_n)$ algorithm, accelerated with Andersen iterations, to obtain the fixed point of the above equation. 
When $\mathcal A$ is a rank deficient operator, the forward and backward iterations are guaranteed to converge when the Lipshitz constant of the network $H=(F-I)$ is less than unity and the step size $\gamma < 1/2$. 
With this setting, DEQ schemes compute the gradient of $F$ using Jacobian iterations. This approach eliminates the need for algorithm unrolling, thus quite significantly reducing the memory demand of the algorithm; a single data consistency block and deep learning block $F$ is sufficient to implement the algorithm. While other algorithms such as ADMM and proximal gradients can be used \cite{gilton2021deep,ryu2019plug}, the convergence guarantees are not valid when $\mathcal A$ is rank deficient. 

\vspace{-1.25em}\subsection{Monotone operator learning and proposed algorithm}
To overcome the above challenges, we focus on choosing $F$ in \eqref{fixedpoint} as a monotone operator \cite{winston2020monotone}. An operator is $m$-monotone if it satisfies the condition:

\begin{figure}[t!]
	\vspace{-1em}
	\includegraphics[width=0.5\textwidth]{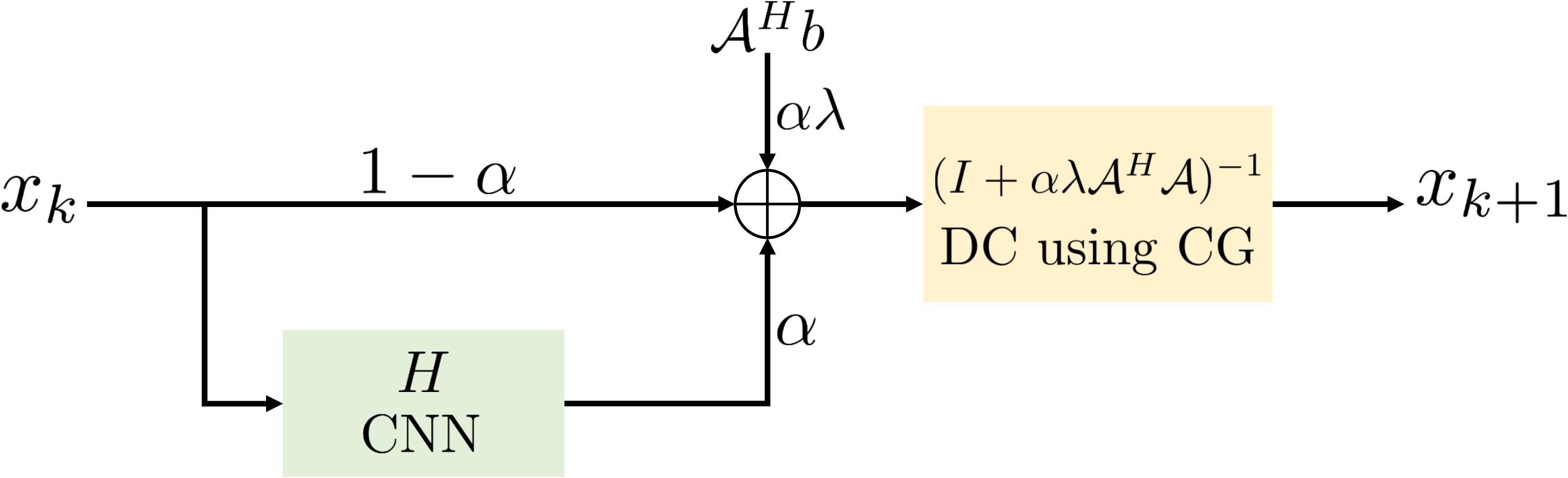}
	\vspace{-1em}
	\label{architecture}
	\caption{Architecture of the iterative block in monotone operator learning \eqref{fwdbwd}. We iterate the above block to convergence, which is guaranteed if the Lipshitz constant of $H$ is lower than $1-m$ and $\alpha<2m/(2-m)^2$. The Lipshitz constraint is enforced by the regularization term \eqref{lipest} during training. }\vspace{-1.5em}
\end{figure}

\vspace{-1em}
\begin{equation}\label{monotone}
\Big\langle (x-y)~,F(x)-F(y)\Big\rangle > m \|x-y\|_2^2
\end{equation}
\vspace{-1em}

We note that the differential $F=\partial \varphi$ is monotone if $\varphi$ is convex. Although the converse is not true in general, DEQ schemes with monotone operator can guarantee convergence and robustness, as shown below. 
We focus on forward backward splitting, where we rewrite \eqref{fixedpoint} as $I+\alpha G-I+\alpha F=0$, for some $\alpha >0$. This simplifies to the 
fixed point iteration: 

\vspace{-1em}
\begin{eqnarray}\label{fwdbwdgeneral}
f_{k+1} &=& (I+\alpha G)^{-1} \Big(f_k - \alpha F(f_k)\Big)
\end{eqnarray}
\vspace{-1em}

with the following convergence guarantee:

\begin{prop}\cite{ryu2019plug}.
	The forward backward optimization algorithm will converge to the fixed point of \eqref{fixedpoint} if  $\alpha <2m/L^2$, where $F$ is $m$-monotone with a Lipshitz constant of $L$.
\end{prop}

\begin{figure*}[t!]
	\subfigure[Lipshitz]{\includegraphics[width=0.24\textwidth]{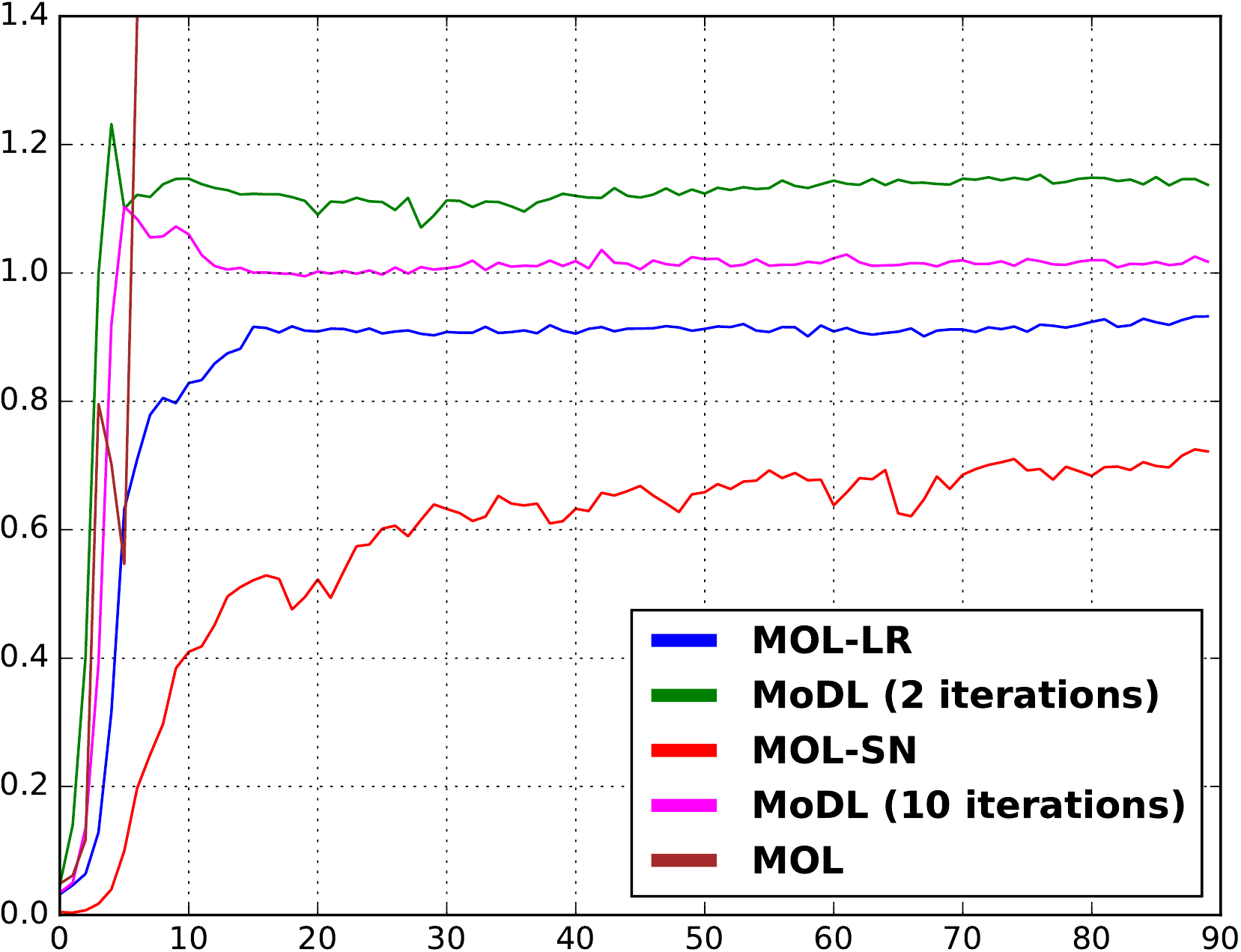}}
	\subfigure[nFE]{\includegraphics[width=0.24\textwidth]{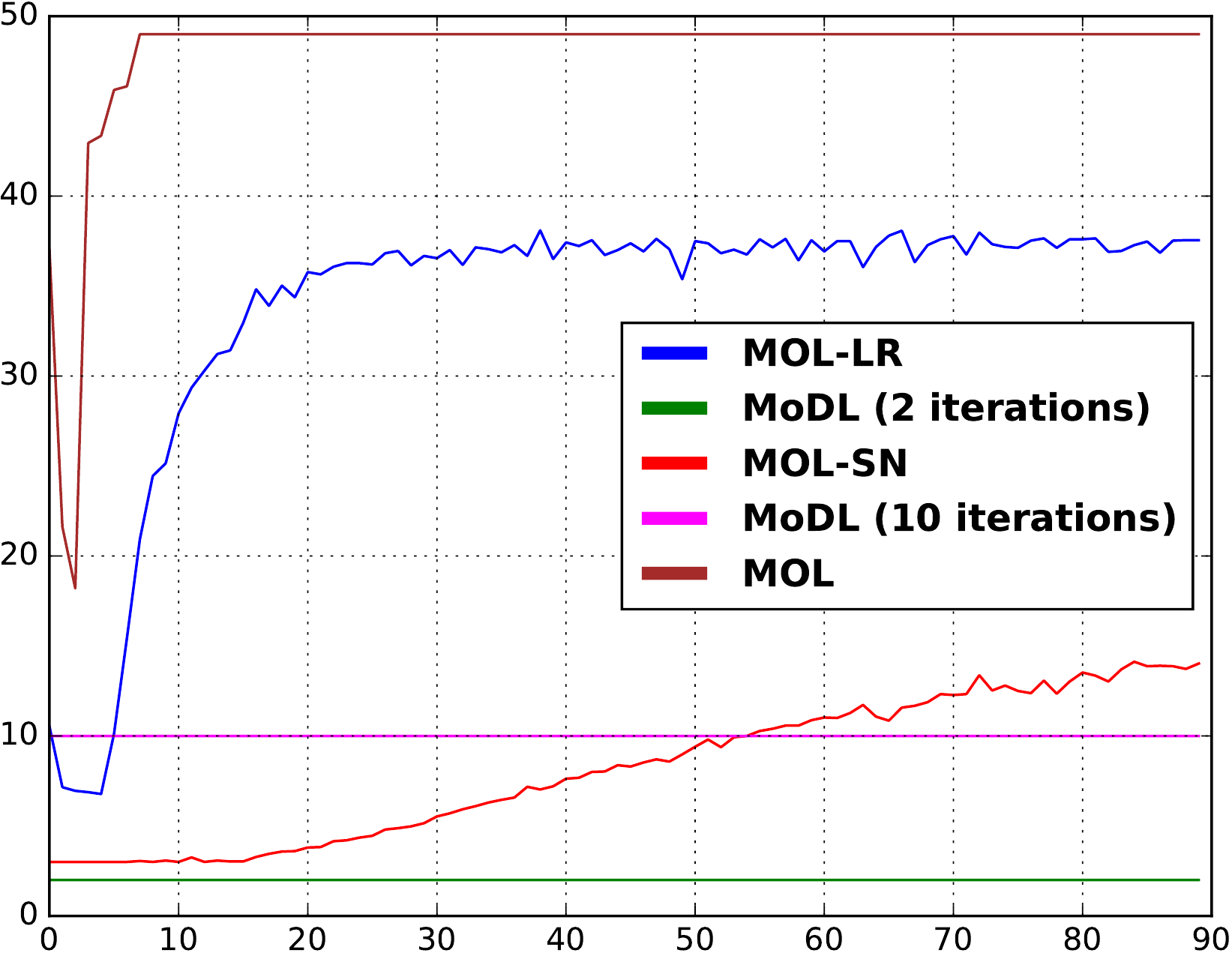}}	
	\subfigure[Training error]{\includegraphics[width=0.24\textwidth]{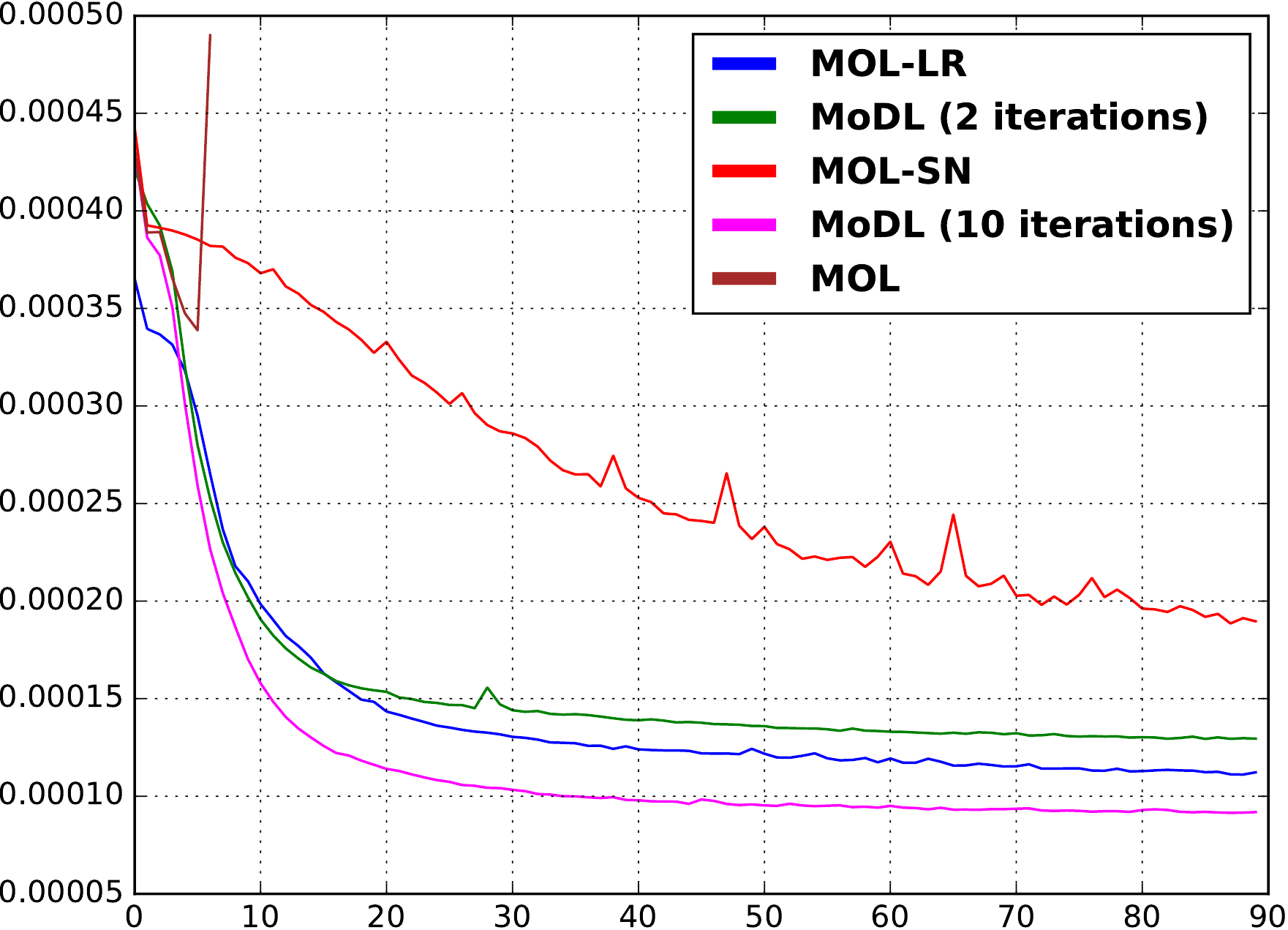}}
	\subfigure[Validation error]{\includegraphics[width=0.24\textwidth]{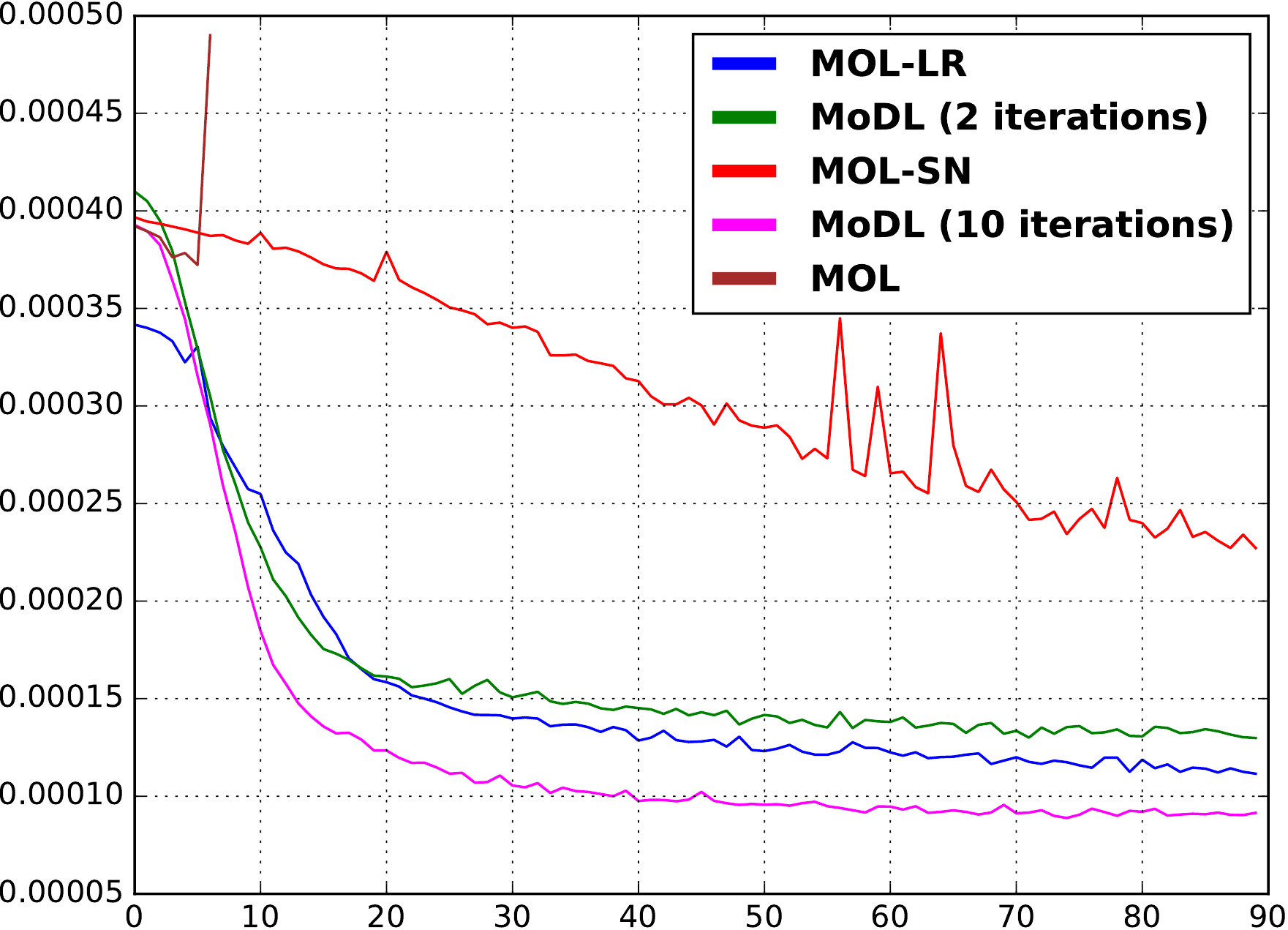}}\vspace{-1.0em}
	\caption{Comparison of MOL algorithms: (a) shows the evolution of the estimated local Lipshitz constant of $H$ in \eqref{lipest}. (b) shows the number of forward iterations. We note that as the Lipshitz constant increases, more number of forward iterations are required for convergence. We plot the training and validation errors of the models in (c) and (d), respectively. The results show that the proposed Lipshitz regularized MOL (MOL-LR) is able to offer comparable results to unrolled schemes. Spectral normalization (MOL-SN) can offer convergence, but the low representation power of $H$ translates to higher errors.   }
	\label{deqevolution}\vspace{-1.0em}
\end{figure*}

\vspace{-1.5em}\subsection{Residual network is a monotone operator}
We note that \cite{winston2020monotone} relies on a specially designed network layer that guarantee that $F$ is monotone. Unfortunately, such single layer networks have low representation power to be used in our setting. We instead show that a residual network $F=I-H$ is $m$-monotone, under the following condition.
\begin{prop}
	$F$ is an $m$-monotone network if the Lipshitz constant of $H$ is less than $1-m$. The Lipshitz constant of $F$ is bounded by $(2-m)$.
\end{prop}
We hence substitute for $F$ in \eqref{fwdbwdgeneral} to obtain
\begin{eqnarray}\label{fwdbwd}
x_{k+1} &=& Q^{-1}\Big(1-\alpha)x_k + \alpha H(x_k)+ \alpha \lambda\, \mathcal A^H b\Big), 
\end{eqnarray}
where $Q =  (I+\alpha \lambda\, \mathcal A^H\mathcal A)$.
We thus obtain the result:
\begin{prop}
	The iterative algorithm specified by \eqref{fwdbwd} will converge to the fixed point of \eqref{fixedpoint}, denoted by $f^*(b)$ if $\alpha < 2m/(2-m)^2$. 
\end{prop}
We note that when $\alpha=1$, the algorithm simplifies to the iterative approach in \cite{aggarwal2018modl}; this algorithm will converge in a DEQ setting  when $m>0.76$ or when the Lipshitz constant of $H$ is less than $0.24$.
\vspace{-1em}\subsection{Robustness to input perturbations}
A challenge with deep learning networks is their vulnerability to adversarial perturbations. The use of monotone operators allows us to bound the norm of the perturbations in the output in terms of the perturbations in the measurements. 
\begin{prop}
Let $f^*(a)$ and $f^*(b)$ be the fixed points corresponding to two set of measurements, denoted by $a$ and $b$, respectively. We then have 
\begin{equation}\label{bound}
\|f^{*}(a)-f^{*}(b)\|_2 \leq  ~~\frac{\alpha \lambda}{1-\sqrt{\left( 1-2\alpha m + \alpha^2 (2-m)^2\right)}} \|a-b\|_2
\end{equation}
\end{prop}
When $\alpha \rightarrow 0$, we have $\|f^{*}(a)-f^{*}(b)\|_2 \leq \lambda \|a-b\|_2/m$. This result shows that $m$ plays a very important role in the robustness of the system. In particular, higher Lipshitz constant of $H$ implies networks with improved representation power, and hence can potentially offer higher performance. However, the improved performance will come with a smaller $m$ and hence higher sensitivity to input perturbations.  
\vspace{-1em}\subsection{Local Lipshitz regularization}
The popular approach for constraining the Lipshitz bound of a network is spectral normalization of each layer. However, the product of the spectral bounds results in a very conservative estimate for the Lipshitz constant of the network; using spectral normalization results in networks with poor representation power, and hence worse results than unrolled networks that do not require these constraints. 


\begin{figure*}[t!]
	\begin{center}
	\subfigure[Ground Truth]{\includegraphics[width=0.18\textwidth,trim={5cm 3cm 2cm 5.5cm},clip]{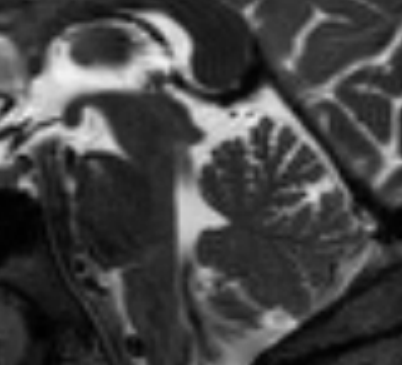}}
	\subfigure[SENSE, 34.02]{\includegraphics[width=0.18\textwidth,trim={5cm 3cm 2cm 5.5cm},clip]{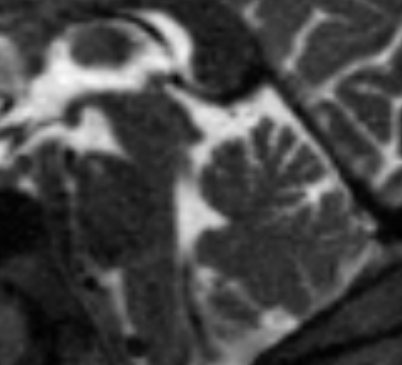}}
	\subfigure[MOL-SN, 35.22]{\includegraphics[width=0.18\textwidth,trim={5cm 3cm 2cm 5.5cm},clip]{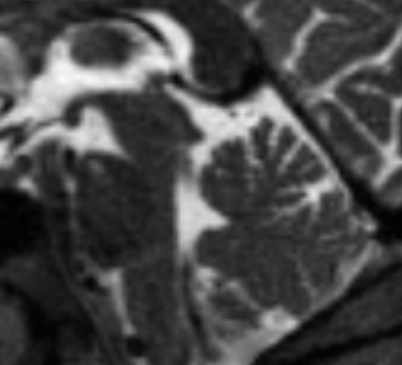}}
	\subfigure[MoDL, 37.86]{\includegraphics[width=0.18\textwidth,trim={5cm 3cm 2cm 5.5cm},clip]{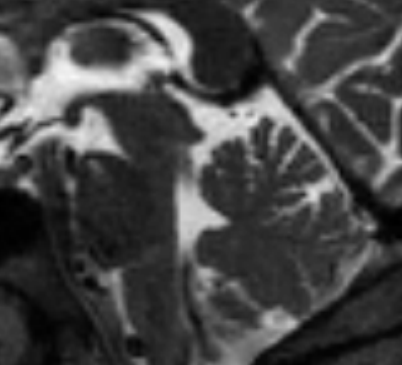}}
	\subfigure[\textbf{MOL-LR, 37.53}]{\includegraphics[width=0.18\textwidth,trim={5cm 3cm 2cm 5.5cm},clip]{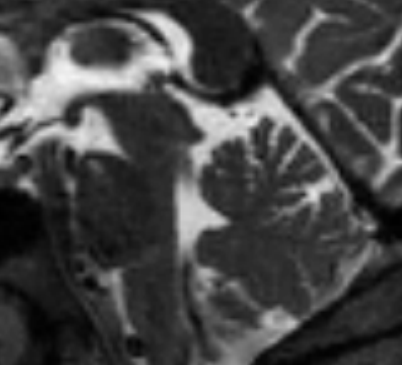}}\\
	\vspace{-1em}
	\subfigure[Mask]{\includegraphics[width=0.18\textwidth]{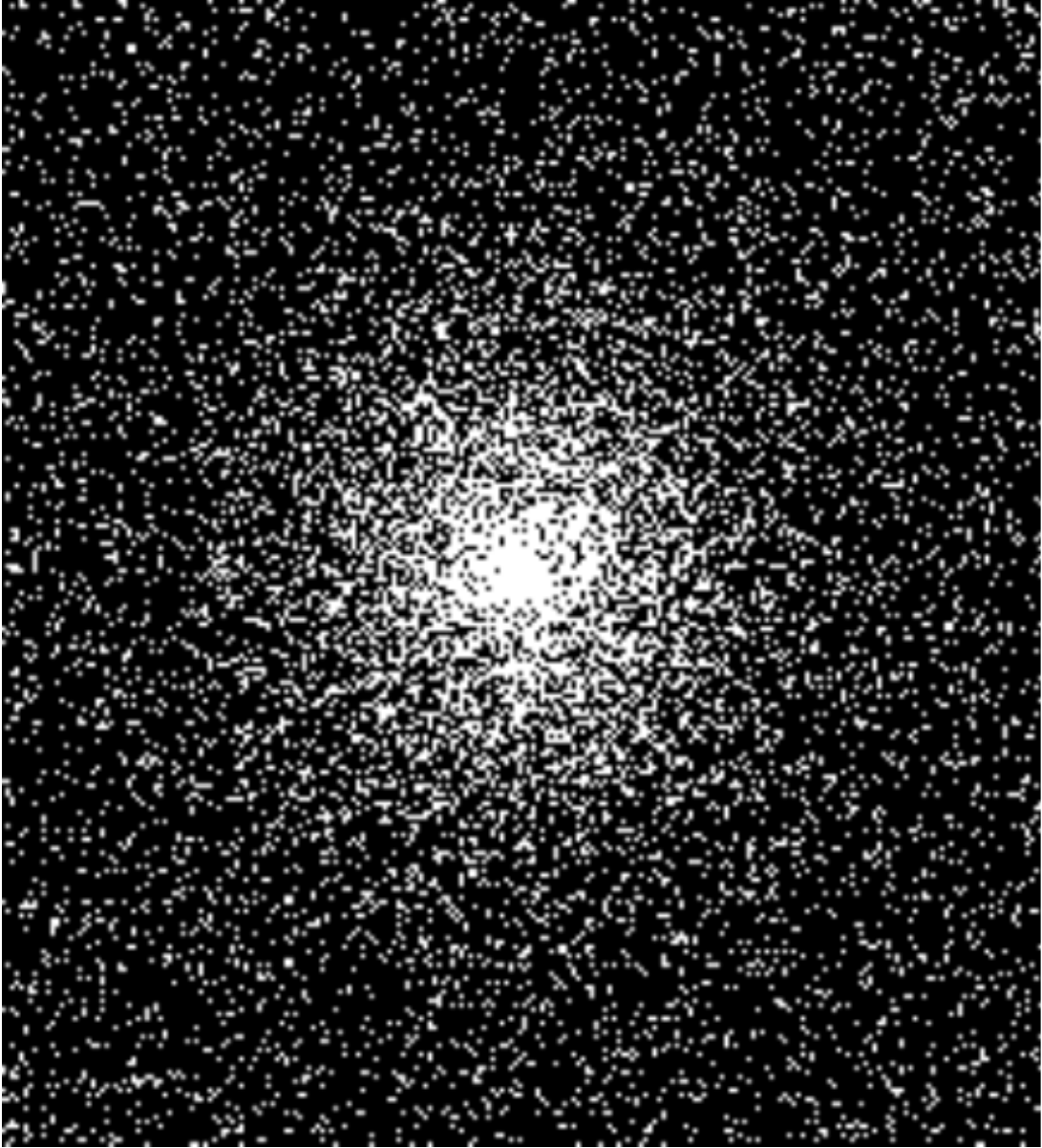}}
	\subfigure[SENSE]{\includegraphics[width=0.18\textwidth]{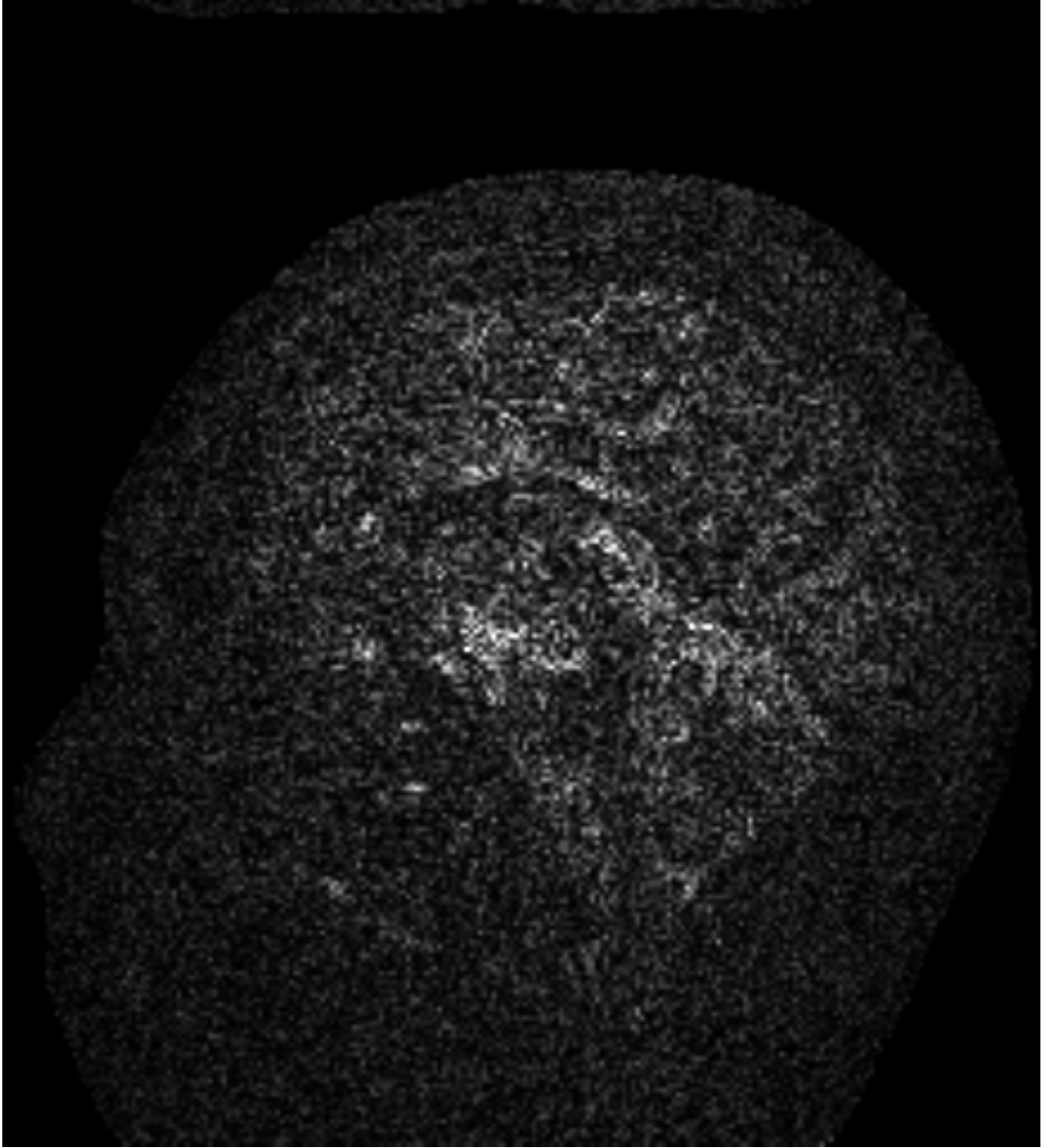}}
	\subfigure[MOL-SN]{\includegraphics[width=0.18\textwidth]{mds_err.pdf}}
	\subfigure[MoDL (10 unrolls)]{\includegraphics[width=0.18\textwidth]{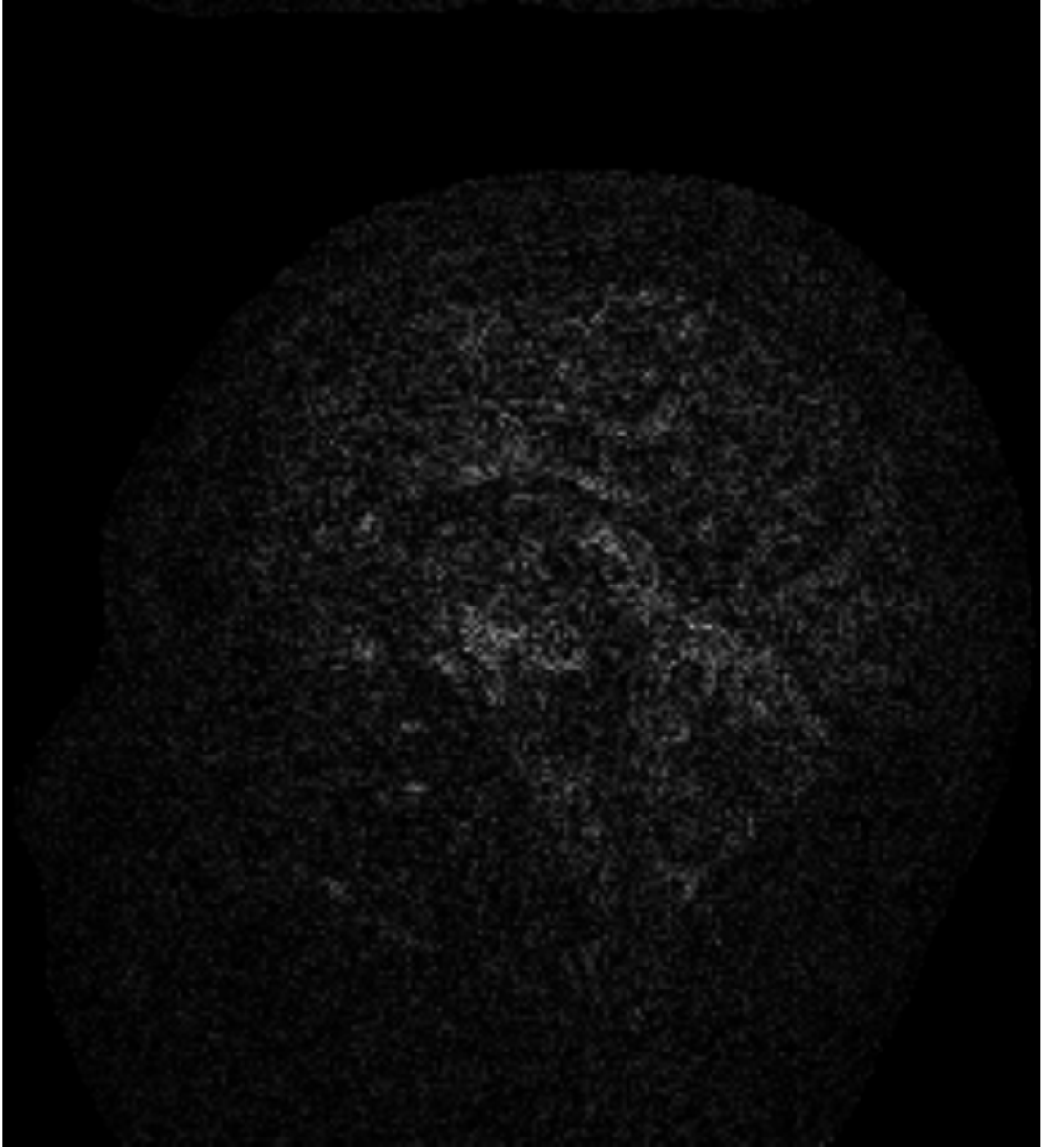}}
	\subfigure[\textbf{MOL-LR}]{\includegraphics[width=0.18\textwidth]{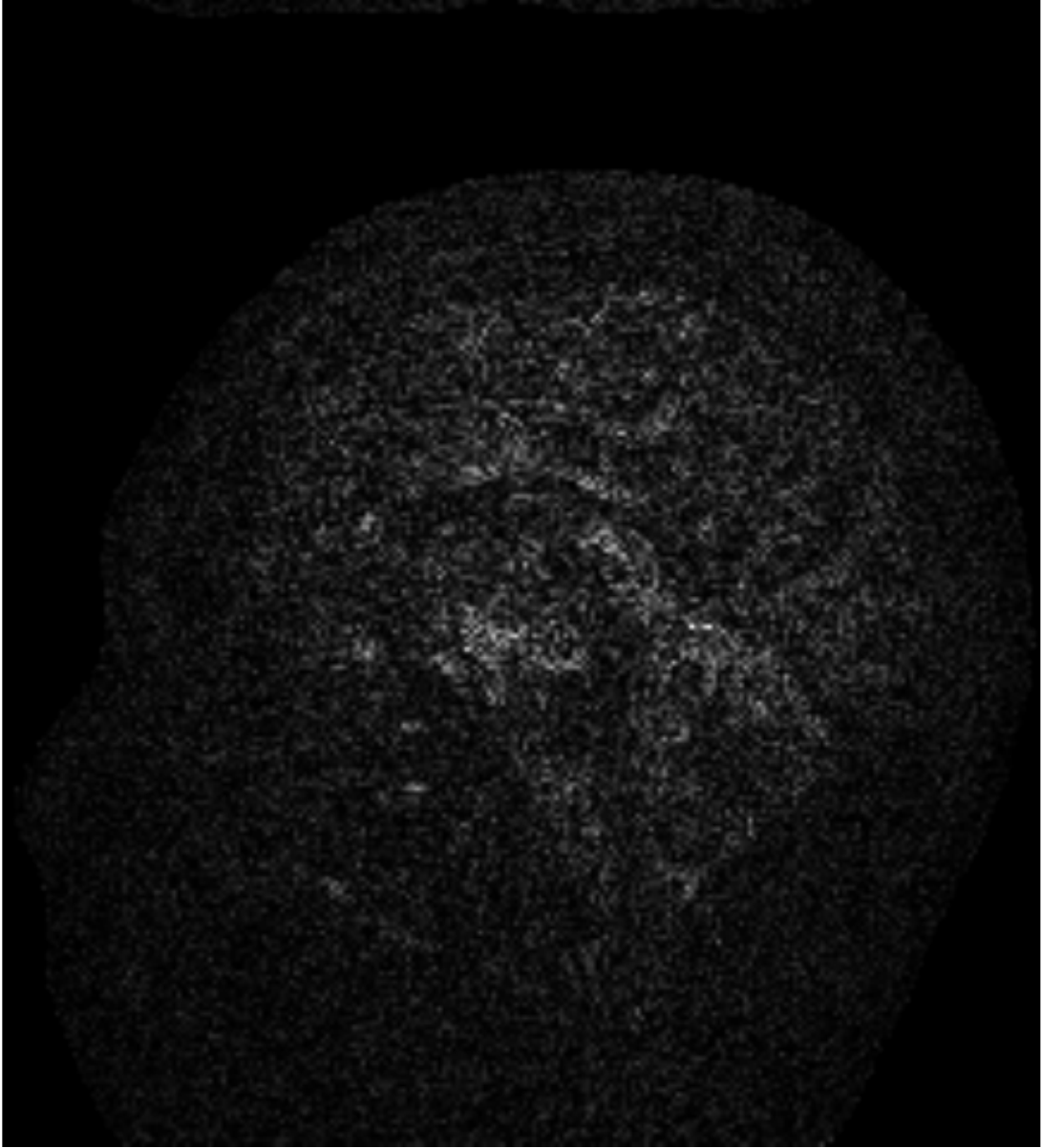}}	\vspace{-0.6em}

	\end{center}\vspace{-1.3em}
	\caption{Reconstruction results of 6-fold accelerated brain data. Top row are zoomed sections of the reconstructions. The proposed monotone operator learning with Lipshitz regularization (MOL-LR) is compared against SENSE, unrolled MoDL with 10 iterations and MOL-SN reconstructions. MOL-LR is a Lipschitz regularized network to ensure convergence while MOL-SN performs spectral normalization of the network weights. The top row displays magnitude of coil sensitivity weighted images and the bottom row shows corresponding error images. The PSNR in dB is reported for the slice in each case. Proposed MOL-LR is highlighted in bold. The GPU memory demand of MOL-LR is ten times lower than that of MODL with 10 unrolling steps, while its performance is comparable.}
	\label{6x}\vspace{-1.5em}
\end{figure*}
In this work, we propose to use a regularization term to constrain the Lipshitz constant of $H$.  Instead of global Lipshitz constants over the entire domain of $H$, we focus on evaluating the local Lipshitz constants to the subregion of the domain where the images live. Specifically, for each batch, we solve an optimization problem involving an adversarial perturbation of the fixed point:
\begin{equation}\label{lipest}
l(f^*) = \max_{\epsilon}\frac{\|H(f^*+\epsilon)-H(f)\|_2^2}{\|\epsilon\|_2^2}
\end{equation}
We optimize the above expression using a few gradient ascent steps during each epoch. 
The total loss at each epoch is the sum of the MSE loss obtained by comparing the reconstructions to the true images and the estimated Lipshitz constants of $H$. 
\begin{equation}\label{key}
\mathcal {L} = \|f^*(\mathcal A^H b) - f_{\rm orig}\|_2^2 +  l(f^*)
\end{equation}





\begin{table}[b!]\vspace{-1.5em}
\fontsize{8}{16}
\selectfont
\centering
\renewcommand{\arraystretch}{0.7}
\begin{tabular}{|c|cc|cc|}
\hline
 & \multicolumn{2}{|c|}{4x} & \multicolumn{2}{|c|}{6x} \\ 
Methods & PSNR (dB) & SSIM & PSNR (dB) & SSIM \\ \hline 
SENSE &37.39 &0.983 &34.04 &0.967\\
MoDL &41.02 &0.992 &37.98 &0.985\\
MOL-SN & 38.31 &0.985 &35.19 &0.972 \\
\textbf{MOL-LR} & \textbf{40.75} &\textbf{0.991} & \textbf{37.71} &\textbf{0.984}\\ \hline
\end{tabular}\vspace{-0.5em}
\caption{Quantitative comparison of SENSE, MoDL, MOL-SN and MOL-LR reconstructions. Average Peak Signal to Noise Ratio (PSNR) and Structural Similarity (SSIM) are reported for 30 test slices. }
\label{tab:comp} 
\vspace{-1em}
\end{table}

\section{Experiments and Results}

The experiments were performed on 3D CUBE brain MRI dataset collected from 9 subjects at the University of Iowa using a 12-channel head coil. The dataset was split into 4 subjects for training, 2 for validation and the remaining 3 for testing. We chose the middle slices from each subject. 2D non-uniform variable density undersampling masks with different acceleration factors were used for the experiments. We used a five layer deep network as in \cite{aggarwal2018modl} to implement $H$. The parameter $\alpha$ was chosen as $2m/(2-m)^2$ to guarantee convergence. We did not use batch normalization because it is difficult to restrict the Lipshitz constant of $H$. 

We characterize the MOL framework in Fig. \ref{deqevolution}. It is observed from (a)-(e) that the MOL approach without spectral normalization or Lipshitz regularization (red curve) diverges rapidly. In particular, the forward  iterations (nFE) increase to the maximum without converging, when back-propagation using Jacobian iterations provide erroneous gradients. It is observed that the forward iterations in the MOL approach with spectral normalization of $H$ (MOL-SN) converges rapidly, as seen from the low number of forward iterations (nFE). However, this approach results in higher training and testing errors than the unrolled MoDL. This can be attributed to the low local Lipshitz constants (hence low representation power) of the $H$ network. By contrast, the Lipshitz regularization (MOL-LR) allows us to keep the Lipshitz constant around 0.9, which offers higher representation power and image quality comparable to MoDL with 2 iterations. The 10 iteration MoDL scheme does not require any constraints on the networks and hence can offer marginally higher performance. However, its Lipshitz constant is higher than unity and hence we dont have any guarantees on robustness as in \eqref{bound}. In addition, the memory footprint of MOL is ten times smaller.

Comparison of the methods are shown in Fig. \ref{6x} and Table \ref{tab:comp} respectively. Results show that the proposed DEQ scheme is able to offer similar performance compared to MoDL with 10 iterations. The memory footprint of MOL-LR is around ten times smaller than MoDL, making it an attractive option in higher dimensional problems. The MoDL scheme cannot be directly applied in 3D/4D setting due to its high memory demand.

\vspace{-1.5em}
\section{Conclusion}
\vspace{-1em}

We introduced a monotone operator learning framework for improved model-based deep learning of inverse problems. The proposed framework comes with guaranteed convergence and robustness. We introduced a Lipshitz regularization scheme to control the monotonicity parameter, which offers a trade-off between performance and robustness. The application of the framework to parallel MRI reconstruction setting shows that the performance of the monotone operator learning framework can match the quality of unrolled approaches, albeit with around 10 fold reduction in memory requirement. This work will enable the extension of MoDL schemes to multi-dimensional problems.

\vspace{-1em}
\section{Compliance with Ethical Standards}
\vspace{-1em}

This research study was conducted using human subject data. Approval was granted by the Ethics Committee of the institute where the data were acquired.
\vspace{-1em}

\bibliographystyle{IEEEbib}
\bibliography{refs}
\vspace{-2em}
\end{document}